\documentclass{article}

\usepackage{amsmath}
\usepackage{amsfonts}
\usepackage{CJKutf8} %
\usepackage[usenames, dvipsnames]{color}
\usepackage{booktabs}
\usepackage{fancyhdr}
\usepackage{graphicx}
\usepackage{hyperref}
\usepackage{multirow}
\usepackage{pgfplots}
\usepackage{skak} %
\usepackage{tabularx}
\usepackage[thinlines]{easytable} %
\usepackage{tikz}
\usepackage{times}
\usepackage[shortlabels]{enumitem}
\usepackage{placeins}
\usepackage[title]{appendix}

\usetikzlibrary{arrows.meta}

\definecolor{darkgreen}{RGB}{0, 125, 0}
\definecolor{orange}{RGB}{255, 125, 125}

\newcommand{\muzero}{\emph{MuZero}}
\newcommand{\reanalyze}{\emph{MuZero Reanalyze}}
\newcommand{\alphazero}{\emph{AlphaZero}}
\newcommand{\alphagozero}{\emph{AlphaGo Zero}}

\newcommand{\stockfish}{\emph{Stockfish}}
\newcommand{\elmo}{\emph{Elmo}}

\newcommand{\expect}[1]{\mathbb{E}\left[ {#1} \right]}

\definecolor{colorwin}{RGB}{156,255,161}
\definecolor{colordraw}{RGB}{220,220,255}
\definecolor{colorloss}{RGB}{255,161,156}

\title{Mastering Atari, Go, Chess and Shogi by Planning with a Learned Model}

\author{
Julian Schrittwieser,$^{1\ast}$ Ioannis Antonoglou,$^{1,2\ast}$ Thomas Hubert,$^{1\ast}$ \\
Karen Simonyan,$^{1}$ Laurent Sifre,$^{1}$ Simon Schmitt,$^{1}$ Arthur Guez,$^{1}$ \\
Edward Lockhart,$^{1}$ Demis Hassabis,$^{1}$ Thore Graepel,$^{1,2}$ Timothy Lillicrap,$^{1}$ \\
David Silver$^{1,2\ast}$
\\
\\
\normalsize{$^{1}$DeepMind, 6 Pancras Square, London N1C 4AG.}
\\
\normalsize{$^{2}$University College London, Gower Street, London WC1E 6BT.}
\\
\normalsize{$^\ast$These authors contributed equally to this work.}
}

\date{}

\begin{document}

\maketitle

\begin{abstract}
\normalsize
Constructing agents with planning capabilities has long been one of the main challenges in the pursuit of artificial  intelligence. Tree-based planning methods have enjoyed huge success in challenging domains, such as chess and Go, where a perfect simulator is available. However, in real-world problems the dynamics governing the environment are often complex and unknown.
In this work we present the \muzero{} algorithm which, by combining a tree-based search with a learned model, achieves superhuman performance in a range of challenging and visually complex domains, without any knowledge of their underlying dynamics. \muzero{} learns a model that, when applied iteratively, predicts the quantities most directly relevant to planning: the reward, the action-selection policy, and the value function. When evaluated on 57 different Atari games - the canonical video game environment for testing AI techniques, in which model-based planning approaches have historically struggled - our new algorithm achieved a new state of the art. When evaluated on Go, chess and shogi, without any knowledge of the game rules, \muzero{} matched the superhuman performance of the \alphazero{} algorithm that was supplied with the game rules.
\end{abstract}

\section{Introduction}

Planning algorithms based on lookahead search have achieved remarkable successes in artificial intelligence. Human world champions have been defeated in classic games such as checkers \cite{schaeffer:chinook}, chess \cite{campbell:deep-blue}, Go \cite{Silver16AG} and poker \cite{brown2018superhuman, deepstack}, and planning algorithms have had real-world impact in applications from logistics \cite{vlahavas2013planning} to chemical synthesis \cite{alphachem}.
However, these planning algorithms all rely on knowledge of the environment's dynamics, such as the rules of the game or an accurate simulator, preventing their direct application to real-world domains like robotics, industrial control, or intelligent assistants.

Model-based reinforcement learning (RL) \cite{sutton:book} aims to address this issue by first learning a model of the environment's dynamics, and then planning with respect to the learned model. Typically, these models have either focused on reconstructing the true environmental state \cite{pilco:deisenroth,heess:stochastic_value_gradients,levine:learning_guided_policy}, or the sequence of full observations \cite{hafner:planet, kaiser:simple}. However, prior work \cite{state_space_models, hafner:planet, kaiser:simple} remains far from the state of the art in visually rich domains, such as Atari 2600 games \cite{ALE}.
Instead, the most successful methods are based on model-free RL \cite{impala, r2d2, apex} -- i.e. they estimate the optimal policy and/or value function directly from interactions with the environment. However, model-free algorithms are in turn far from the state of the art in domains that require precise and sophisticated lookahead, such as chess and Go.

In this paper, we introduce \muzero{}, a new approach to model-based RL that achieves state-of-the-art performance in Atari 2600, a visually complex set of domains, while maintaining superhuman performance in precision planning tasks such as chess, shogi and Go.
\muzero{} builds upon \alphazero{}'s \cite{Silver18AZ} powerful search and search-based policy iteration algorithms, but incorporates a learned model into the training procedure. \muzero{} also extends \alphazero{} to a broader set of environments including single agent domains and non-zero rewards at intermediate time-steps.

The main idea of the algorithm (summarized in Figure 1) is to predict those aspects of the future that are directly relevant for planning. The model receives the observation (e.g. an image of the Go board or the Atari screen) as an input and transforms it into a hidden state. The hidden state is then updated iteratively by a recurrent process that receives the previous hidden state and a hypothetical next action. At every one of these steps the model predicts the policy (e.g. the move to play), value function (e.g. the predicted winner), and immediate reward (e.g. the points scored by playing a move). The model is trained end-to-end, with the sole objective of accurately estimating these three important quantities, so as to match the improved estimates of policy and value generated by search as well as the observed reward. There is no direct constraint or requirement for the hidden state to capture all information necessary to reconstruct the original observation, drastically reducing the amount of information the model has to maintain and predict; nor is there any requirement for the hidden state to match the unknown, true state of the environment; nor any other constraints on the semantics of state. Instead, the hidden states are free to represent state in whatever way is relevant to predicting current and future values and policies. Intuitively, the agent can invent, internally, the rules or dynamics that lead to most accurate planning.

\section{Prior Work}

Reinforcement learning may be subdivided into two principal categories: model-based, and model-free \cite{sutton:book}. Model-based RL constructs, as an intermediate step, a model of the environment. Classically, this model is represented by a Markov-decision process (MDP) \cite{puterman:MDP} consisting of two components: a state transition model, predicting the next state, and a reward model, predicting the expected reward during that transition. The model is typically conditioned on the selected action, or a temporally abstract behavior such as an option \cite{sutton:between}. Once a model has been constructed, it is straightforward to apply MDP planning algorithms, such as value iteration \cite{puterman:MDP} or Monte-Carlo tree search (MCTS) \cite{coulom:mcts}, to compute the optimal value or optimal policy for the MDP. In large or partially observed environments, the algorithm must first construct the state representation that the model should predict. This tripartite separation between representation learning, model learning, and planning is potentially problematic since the agent is not able to optimize its representation or model for the purpose of effective planning, so that, for example modeling errors may compound during planning.

A common approach to model-based RL focuses on directly modeling the observation stream at the pixel-level. It has been hypothesized that deep, stochastic models may mitigate the problems of compounding error \cite{hafner:planet, kaiser:simple}. However, planning at pixel-level granularity is not computationally tractable in large scale problems. Other methods build a latent state-space model that is sufficient to reconstruct the observation stream at pixel level \cite{wahlstrom:pixels_to_torques,watter:embed_to_control}, or to predict its future latent states \cite{ha:world_model,gelada:deepmdp}, which facilitates more efficient planning but still focuses the majority of the model capacity on potentially irrelevant detail. None of these prior methods has constructed a model that facilitates effective planning in visually complex domains such as Atari; results lag behind well-tuned, model-free methods, even in terms of data efficiency \cite{hado:replay}.

A quite different approach to model-based RL has recently been developed, focused end-to-end on predicting the value function \cite{silver:predictron}. The main idea of these methods is to construct an abstract MDP model such that planning in the abstract MDP is equivalent to planning in the real environment. This equivalence is achieved by ensuring \emph{value equivalence}, i.e. that, starting from the same real state, the cumulative reward of a trajectory through the abstract MDP matches the cumulative reward of a trajectory in the real environment.

The predictron \cite{silver:predictron} first introduced value equivalent models for predicting value (without actions). Although the underlying model still takes the form of an MDP, there is no requirement for its transition model to match real states in the environment. Instead the MDP model is viewed as a hidden layer of a deep neural network. The unrolled MDP is trained such that the expected cumulative sum of rewards matches the expected value with respect to the real environment, e.g. by temporal-difference learning.

Value equivalent models were subsequently extended to optimising value (with actions). TreeQN \cite{farquhar:treeqn} learns an abstract MDP model, such that a tree search over that model (represented by a tree-structured neural network) approximates the optimal value function. Value iteration networks \cite{aviv:vin} learn a local MDP model, such that value iteration over that model (represented by a convolutional neural network) approximates the optimal value function. Value prediction networks \cite{oh:vpn} are perhaps the closest precursor to \muzero{}: they learn an MDP model grounded in real actions; the unrolled MDP is trained such that the cumulative sum of rewards, conditioned on the actual sequence of actions generated by a simple lookahead search, matches the real environment. Unlike \muzero{} there is no policy prediction, and the search only utilizes value prediction.

\section{\muzero{} Algorithm}

\begin{figure}
\includegraphics[width=\textwidth]{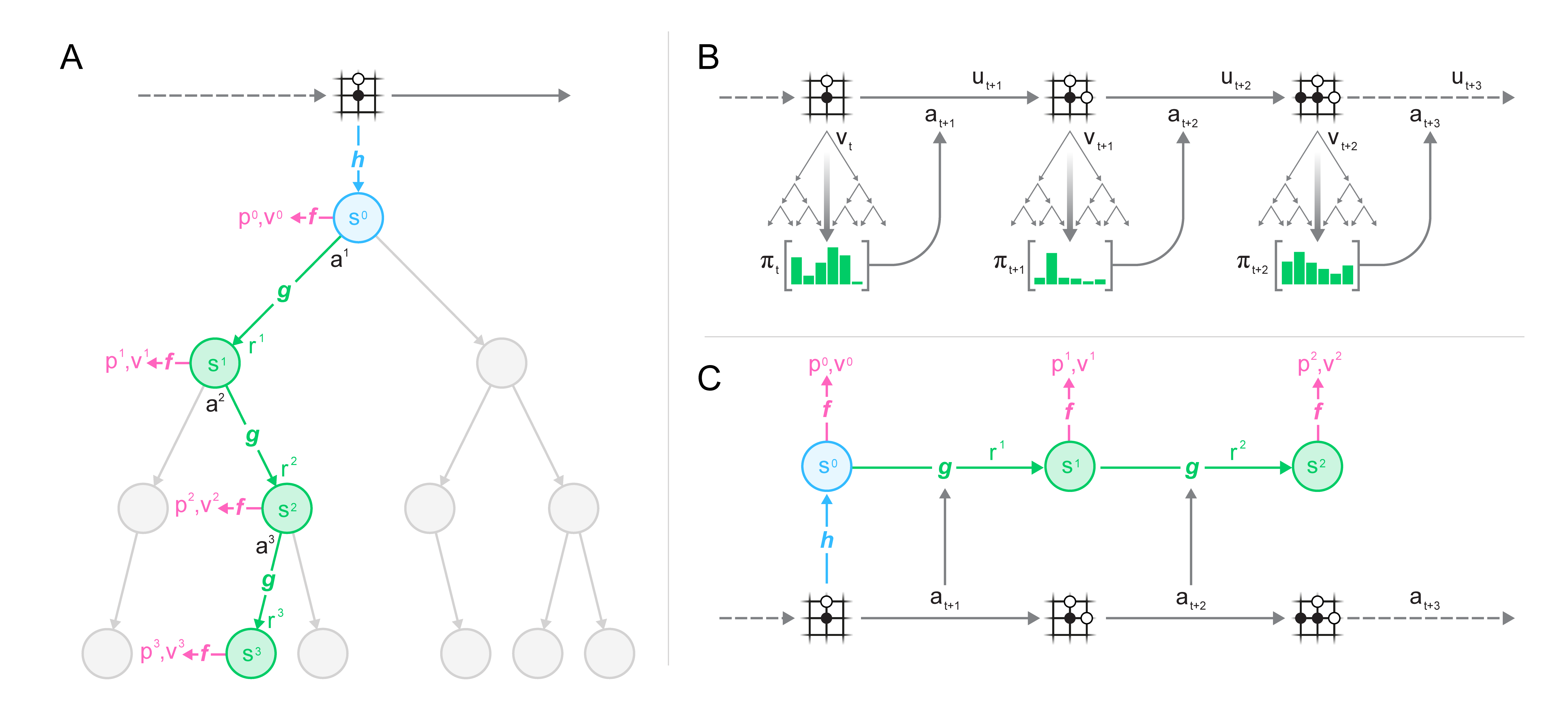}
\caption{
\label{fig:recurrent_search}
\textbf{Planning, acting, and training with a learned model.}
\textbf{(A)} How \muzero{} uses its model to plan. The model consists of three connected components for representation, dynamics and prediction. %
Given a previous hidden state $s^{k-1}$ and a candidate action $a^{k}$, the \emph{dynamics} function $g$ produces an immediate reward $r^{k}$ and a new hidden state $s^{k}$. The policy $p^k$ and value function $v^k$ are computed from the hidden state $s^k$ by a \emph{prediction} function $f$. The initial hidden state $s^0$ is obtained by passing the past observations (e.g. the Go board or Atari screen) into a \emph{representation} function $h$.
\textbf{(B)} How \muzero{} acts in the environment. A Monte-Carlo Tree Search is performed at each timestep $t$, as described in A. An action $a_{t+1}$ is sampled from the search policy $\pi_t$, which is proportional to the visit count for each action from the root node. The environment receives the action and generates a new observation $o_{t+1}$ and reward $u_{t+1}$. At the end of the episode the trajectory data is stored into a replay buffer.
\textbf{(C)} How \muzero{} trains its model. A trajectory is sampled from the replay buffer. For the initial step, the representation function $h$ receives as input the past observations $o_1, ..., o_t$ from the selected trajectory. The model is subsequently unrolled recurrently for $K$ steps. %
At each step $k$, the dynamics function $g$ receives as input the hidden state $s^{k-1}$ from the previous step and the real action $a_{t+k}$.
The parameters of the representation, dynamics and prediction functions are jointly trained, end-to-end by backpropagation-through-time, to predict three quantities: the policy $\mathbf{p}^k \approx \pi_{t+k}$, value function $v^k \approx z_{t+k}$, and reward $r_{t+k} \approx u_{t+k}$, where $z_{t+k}$ is a sample return: either the final reward (board games) or $n$-step return (Atari).
}
\end{figure}

We now describe the \muzero{} algorithm in more detail. Predictions are made at each time-step $t$, for each of $k = 1...K$ steps, by a model $\mu_\theta$, with parameters $\theta$, conditioned on past observations $o_1, ..., o_t$ and future actions $a_{t+1}, ..., a_{t+k}$. The model predicts three future quantities: the policy $\mathbf{p}^k_t \approx \pi(a_{t+k+1} | o_1, ..., o_t, a_{t+1}, ..., a_{t+k})$, the value function $v^k_t \approx \expect{u_{t+k+1} + \gamma u_{t+k+2} + ... | o_1, ..., o_t, a_{t+1}, ..., a_{t+k}}$, and the immediate reward $r^k_t \approx u_{t+k}$, where $u_{.}$ is the true, observed reward, $\pi$ is the policy used to select real actions, and $\gamma$ is the discount function of the environment.

Internally, at each time-step $t$ (subscripts $_t$ suppressed for simplicity), the model is represented by the combination of a \emph{representation} function, a \emph{dynamics} function, and a \emph{prediction} function. The dynamics function, $r^k, s^k = g_\theta(s^{k-1}, a^k)$, is a recurrent process that computes, at each hypothetical step $k$, an immediate reward $r^k$ and an internal state $s^k$. It mirrors the structure of an MDP model that computes the expected reward and state transition for a given state and action \cite{puterman:MDP}. However, unlike traditional approaches to model-based RL \cite{sutton:book}, this internal state $s^k$ has no semantics of environment state attached to it -- it is simply the hidden state of the overall model, and its sole purpose is to accurately predict relevant, future quantities: policies, values, and rewards. In this paper, the \emph{dynamics} function is represented deterministically; the extension to stochastic transitions is left for future work. The policy and value functions are computed from the internal state $s^k$ by the prediction function, $\mathbf{p}^k, v^k = f_\theta(s^k)$, akin to the joint policy and value network of \alphazero{}. The ``root" state $s^0$ is initialized using a representation function that encodes past observations, $s^0 = h_\theta(o_1, ..., o_t)$; again this has no special semantics beyond its support for future predictions.

Given such a model, it is possible to search over hypothetical future trajectories $a^1, ..., a^k$ given past observations $o_1, ..., o_t$. For example, a naive search could simply select the $k$ step action sequence that maximizes the value function. %
More generally, we may apply any MDP planning algorithm to the internal rewards and state space induced by the dynamics function. Specifically, we use an MCTS algorithm similar to \alphazero{}'s search, generalized to allow for single agent domains and intermediate rewards (see Methods). At each internal node, it makes use of the policy, value and reward estimates produced by the current model parameters $\theta$. The MCTS algorithm outputs a recommended policy $\pi_t$ and estimated value $\nu_t$. An action $a_{t+1} \sim \pi_t$ is then selected.

All parameters of the model are trained jointly to accurately match the policy, value, and reward, for every hypothetical step $k$, to corresponding target values observed after $k$ actual time-steps have elapsed. Similarly to \alphazero{}, the improved policy targets are generated by an MCTS search; the first objective is to minimise the error between predicted policy $\mathbf{p}_t^k$ and search policy $\pi_{t+k}$. Also like \alphazero{}, the improved value targets are generated by playing the game or MDP. However, unlike \alphazero{}, we allow for long episodes with discounting and intermediate rewards by \emph{bootstrapping} $n$ steps into the future from the search value, $z_t = u_{t+1} + \gamma u_{t+2} + ... + \gamma^{n-1} u_{t+n} + \gamma^n \nu_{t+n}$. Final outcomes $\{lose, draw, win\}$ in board games are treated as rewards $u_t \in \{ -1, 0, +1 \}$ occuring at the final step of the episode. Specifically, the second objective is to minimize the error between the predicted value $v^k_t$ and the value target, $z_{t+k}$ \footnote{For chess, Go and shogi, the same squared error loss as \alphazero{} is used for rewards and values. A cross-entropy loss was found to be more stable than a squared error when encountering rewards and values of variable scale in Atari. Cross-entropy was used for the policy loss in both cases.}. The reward targets are simply the observed rewards; the third objective is therefore to minimize the error between the predicted reward $r^k_t$ and the observed reward $u_{t+k}$. Finally, an L2 regularization term is also added, leading to the overall loss:

\begin{align}
l_t(\theta) &= \sum_{k=0}^K l^r (u_{t+k}, r_t^k) + l^v(z_{t+k}, v^k_t) + l^p(\pi_{t+k}, \mathbf{p}^k_t) + c ||\theta||^2
\label{muzero_eqn}
\end{align}
where $l^r$, $l^v$, and $l^p$ are loss functions for reward, value and policy respectively.
Supplementary Figure \ref{fig:muzero_equations} summarizes the equations governing how the \muzero{} algorithm plans, acts, and learns.

\section{Results}

\begin{figure} %
\begin{tabularx}{\textwidth}{c c c c}

\textbf{Chess}
& \textbf{Shogi}
& \textbf{Go}
& \textbf{Atari} \\

\newgame
\scalebox{0.62}{\newgame \notationoff \showboard} &
\scalebox{0.745}{
\begin{CJK}{UTF8}{min}
\begin{TAB}(e,0.5cm,0.5cm){|c|c|c|c|c|c|c|c|c|}{|c|c|c|c|c|c|c|c|c|}
\rotatebox[origin=c]{180}{香} & \rotatebox[origin=c]{180}{桂} & \rotatebox[origin=c]{180}{銀} & \rotatebox[origin=c]{180}{金} & \rotatebox[origin=c]{180}{玉} & \rotatebox[origin=c]{180}{金} & \rotatebox[origin=c]{180}{銀} & \rotatebox[origin=c]{180}{桂} & \rotatebox[origin=c]{180}{香}\\
 & \rotatebox[origin=c]{180}{飛} &  &  &  &  &  & \rotatebox[origin=c]{180}{角} & \\
\rotatebox[origin=c]{180}{歩} & \rotatebox[origin=c]{180}{歩} & \rotatebox[origin=c]{180}{歩} & \rotatebox[origin=c]{180}{歩} & \rotatebox[origin=c]{180}{歩} & \rotatebox[origin=c]{180}{歩} & \rotatebox[origin=c]{180}{歩} & \rotatebox[origin=c]{180}{歩} & \rotatebox[origin=c]{180}{歩}\\
 &  &  &  &  &  &  &  & \\
 &  &  &  &  &  &  &  & \\
 &  &  &  &  &  &  &  & \\
歩 & 歩 & 歩 & 歩 & 歩 & 歩 & 歩 & 歩 & 歩\\
 & 角 &  &  &  &  &  & 飛 & \\
香 & 桂 & 銀 & 金 & 玉 & 金 & 銀 & 桂 & 香\\
\end{TAB}
\end{CJK}
}%
&
\scalebox{0.195}{
\begin{tikzpicture}
\draw[help lines] (0,0) grid (18,18);
\draw[fill=gray] (3,3) circle[radius=0.1];
\draw[fill=gray] (3,9) circle[radius=0.1];
\draw[fill=gray] (3,15) circle[radius=0.1];
\draw[fill=gray] (9,3) circle[radius=0.1];
\draw[fill=gray] (9,9) circle[radius=0.1];
\draw[fill=gray] (9,15) circle[radius=0.1];
\draw[fill=gray] (15,3) circle[radius=0.1];
\draw[fill=gray] (15,9) circle[radius=0.1];
\draw[fill=gray] (15,15) circle[radius=0.1];
\end{tikzpicture}
}
&
\includegraphics[width=0.219\textwidth,height=0.219\textwidth]{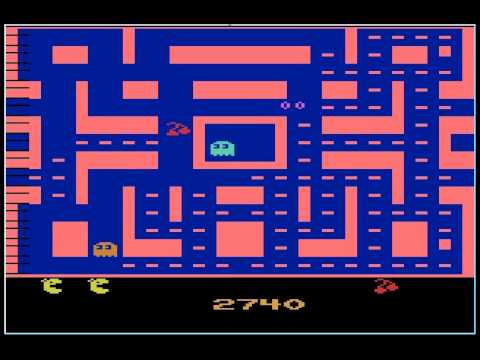}
\\

\multicolumn{4}{c}{
  \hspace{-3.25em}
  \includegraphics[width=\textwidth + 2.6em]{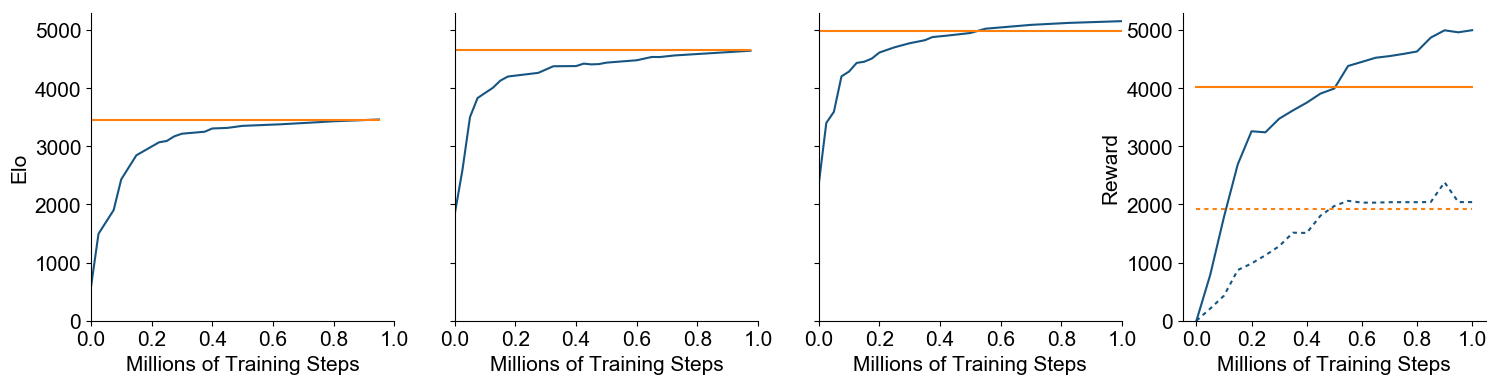}
}

\\

\end{tabularx}

\caption{
\label{fig:results}
\textbf{Evaluation of \muzero{} throughout training in chess, shogi, Go and Atari.} The x-axis shows millions of training steps. For chess, shogi and Go, the y-axis shows Elo rating, established by playing games against \alphazero{} using 800 simulations per move for both players. \muzero{}'s Elo is indicated by the blue line, \alphazero{}'s Elo by the horizontal orange line. For Atari, mean (full line) and median (dashed line) human normalized scores across all 57 games are shown on the y-axis. The scores for R2D2 \cite{r2d2}, (the previous state of the art in this domain, based on model-free RL) are indicated by the horizontal orange lines. Performance in Atari was evaluated using 50 simulations every fourth time-step, and then repeating the chosen action four times, as in prior work \cite{dqn}.
}
\end{figure}
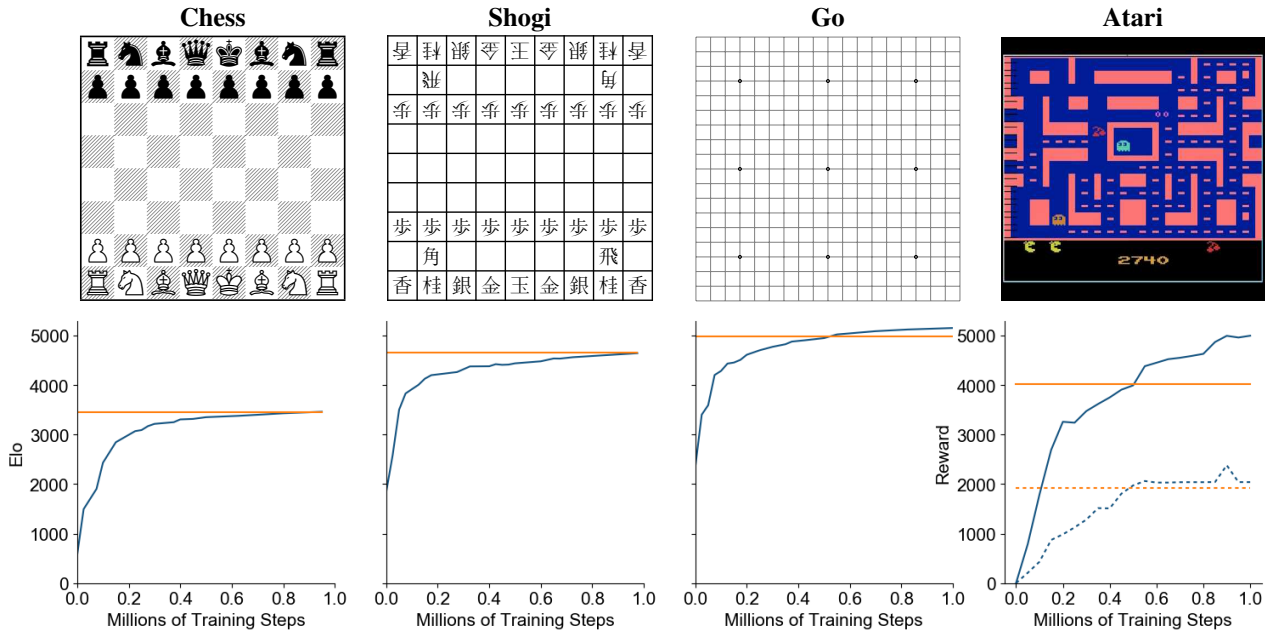 %

We applied the \muzero{} algorithm to the classic board games Go, chess and shogi~\footnote{Imperfect information games such as Poker are not directly addressed by our method.}, as benchmarks for challenging planning problems, and to all 57 games in the Atari Learning Environment \cite{ALE}, as benchmarks for visually complex RL domains.

In each case we trained \muzero{} for $K=5$ hypothetical steps. Training proceeded for 1 million mini-batches of size 2048 in board games and of size 1024 in Atari. During both training and evaluation, \muzero{} used 800 simulations for each search in board games, and 50 simulations for each search in Atari. The representation function uses the same convolutional \cite{convnet} and residual \cite{he:resnet} architecture as \alphazero{}, but with 16 residual blocks instead of 20. The dynamics function uses the same architecture as the representation function and the prediction function uses the same architecture as \alphazero{}. All networks use 256 hidden planes (see Methods for further details).

Figure \ref{fig:results} shows the performance throughout training in each game. In Go, \muzero{} slightly exceeded the performance of \alphazero{}, despite using less computation per node in the search tree (16 residual blocks per evaluation in \muzero{} compared to 20 blocks in \alphazero{}). This suggests that \muzero{} may be caching its computation in the search tree and using each additional application of the dynamics model to gain a deeper understanding of the position.

In Atari, \muzero{} achieved a new state of the art for both mean and median normalized score across the 57 games of the Arcade Learning Environment, outperforming the previous state-of-the-art method R2D2 \cite{r2d2} (a model-free approach) in 42 out of 57 games, and outperforming the previous best model-based approach SimPLe \cite{kaiser:simple} in all games (see Table \ref{tab:atari-results-at30}).

We also evaluated a second version of \muzero{} that was optimised for greater sample efficiency. Specifically, it reanalyzes old trajectories by re-running the MCTS using the latest network parameters to provide fresh targets (see Appendix \ref{reanalyze}). When applied to 57 Atari games, using 200 million frames of experience per game, \reanalyze{} achieved 731\% median normalized score, compared to 192\%, 231\% and 431\% for previous state-of-the-art model-free approaches IMPALA \cite{impala}, Rainbow \cite{rainbow} and LASER \cite{laser} respectively.

\begin{table}

\begin{tabularx}{\textwidth}{c c c c c c}

\toprule
Agent &   Median &     Mean & Env. Frames & Training Time & Training Steps \\
\midrule
Ape-X \cite{apex} &  434.1\% & 1695.6\% & 22.8B & 5 days & 8.64M\\
R2D2 \cite{r2d2} & 1920.6\% & 4024.9\% & 37.5B & 5 days & 2.16M \\
\muzero{} & \textbf{2041.1\%
} & \textbf{4999.2\%
} & 20.0B & 12 hours & 1M \\
\midrule
IMPALA \cite{impala}  & 191.8\% & 957.6\% & 200M & -- & -- \\
Rainbow \cite{rainbow} &  231.1\% & -- & 200M & 10 days & -- \\
UNREAL\textsuperscript{a} \cite{unreal} & 250\%\textsuperscript{a} & 880\%\textsuperscript{a} & 250M & -- & -- \\
LASER \cite{laser} & 431\% & -- & 200M & -- & -- \\
\reanalyze{} & \textbf{731.1\%} & \textbf{2168.9\%} & 200M & 12 hours & 1M \\
\bottomrule
\end{tabularx}

\caption{
\label{tab:atari-comparison}
\textbf{Comparison of \muzero{} against previous agents in Atari}. We compare separately against agents trained in large (top) and small (bottom) data settings; all agents other than \muzero{} used model-free RL techniques. Mean and median scores are given, compared to human testers. The best results are highlighted in \textbf{bold}. \muzero{} sets a new state of the art in both settings. \textsuperscript{a}Hyper-parameters were tuned per game.
}
\end{table}

To understand the role of the model in \muzero{} we also ran several experiments, focusing on the board game of Go and the Atari game of Ms.~Pacman.

\begin{figure} %
\includegraphics[width=0.5\textwidth]{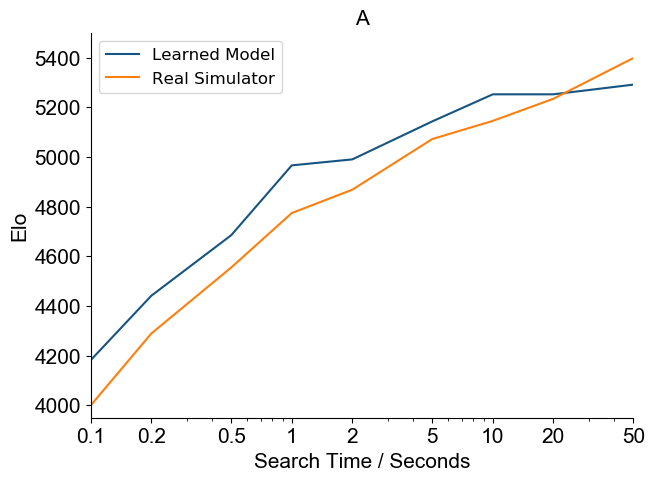}
\includegraphics[width=0.5\textwidth]{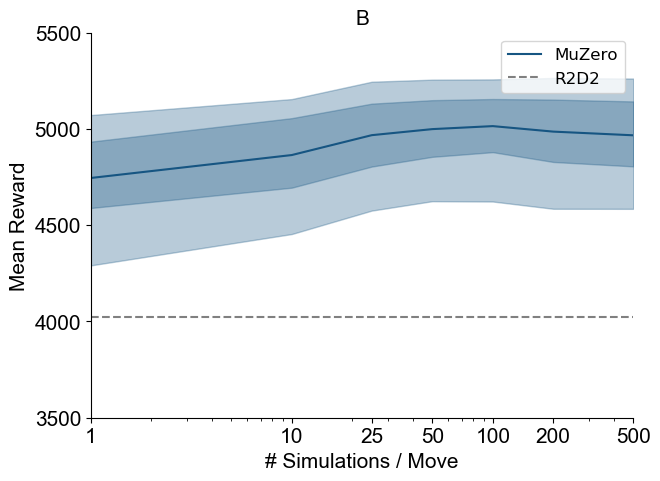}
\includegraphics[width=\textwidth]{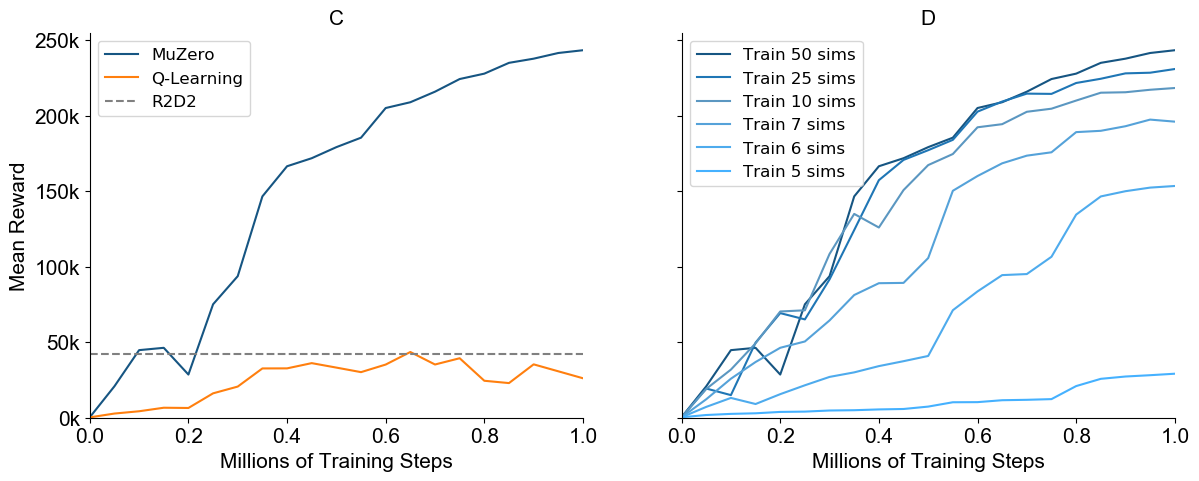}
\caption{
\label{fig:ablations}
\textbf{Evaluations of \muzero{} on Go (A), all 57 Atari Games (B) and Ms.~Pacman (C-D)}. (\textbf{A}) Scaling with search time per move in Go, comparing the learned model with the ground truth simulator. Both networks were trained at 800 simulations per search, equivalent to 0.1 seconds per search. Remarkably, the learned model is able to scale well to up to two orders of magnitude longer searches than seen during training. (\textbf{B}) Scaling of final human normalized mean score in Atari with the number of simulations per search. The network was trained at 50 simulations per search. Dark line indicates mean score, shaded regions indicate 25th to 75th and 5th to 95th percentiles. The learned model's performance increases up to 100 simulations per search. Beyond, even when scaling to much longer searches than during training, the learned model's performance remains stable and only decreases slightly.
This contrasts with the much better scaling in Go (A), presumably due to greater model inaccuracy in Atari than Go.
(\textbf{C}) Comparison of MCTS based training with Q-learning in the \muzero{} framework on Ms.~Pacman, keeping network size and amount of training constant. The state of the art Q-Learning algorithm R2D2 is shown as a baseline. Our Q-Learning implementation reaches the same final score as R2D2, but improves slower and results in much lower final performance compared to MCTS based training. (\textbf{D}) Different networks trained at different numbers of simulations per move, but all evaluated at 50 simulations per move. Networks trained with more simulations per move improve faster, consistent with ablation (B), where the policy improvement is larger when using more simulations per move. Surprisingly, \muzero{} can learn effectively even when training with less simulations per move than are enough to cover all 8 possible actions in Ms.~Pacman.
}
\end{figure}

First, we tested the scalability of planning (Figure \ref{fig:ablations}A), in the canonical planning problem of Go.  We compared the performance of search in \alphazero{}, using a perfect model, to the performance of search in \muzero, using a learned model. Specifically, the fully trained \alphazero{} or \muzero{} was evaluated by comparing MCTS with different thinking times. \muzero{} matched the performance of a perfect model, even when doing much larger searches (up to 10s thinking time) than those from which the model was trained (around 0.1s thinking time, see also Figure \ref{fig:extended_ablations}A).

We also investigated the scalability of planning across all Atari games (see Figure \ref{fig:ablations}B). We compared MCTS with different numbers of simulations, using the fully trained \muzero{}. The improvements due to planning are much less marked than in Go, perhaps because of greater model inaccuracy; performance improved slightly with search time, but plateaued at around 100 simulations. Even with a single simulation -- i.e. when selecting moves solely according to the policy network -- \muzero{} performed well, suggesting that, by the end of training, the raw policy has learned to internalise the benefits of search (see also Figure \ref{fig:extended_ablations}B).

Next, we tested our model-based learning algorithm against a comparable model-free learning algorithm (see Figure \ref{fig:ablations}C). We replaced the training objective of \muzero{} (Equation 1) with a model-free Q-learning objective (as used by R2D2), and the dual value and policy heads with a single head representing the Q-function $Q(\cdot|s_t)$. Subsequently, we trained and evaluated the new model without using any search. When evaluated on Ms.~Pacman, our model-free algorithm achieved identical results to R2D2, but learned significantly slower than \muzero{} and converged to a much lower final score. We conjecture that the search-based policy improvement step of \muzero{} provides a stronger learning signal than the high bias, high variance targets used by Q-learning.

To better understand the nature of \muzero{}'s learning algorithm, we measured how \muzero{}'s training scales with respect to the amount of search it uses \emph{during} training. Figure \ref{fig:ablations}D shows the performance in Ms.~Pacman, using an MCTS of different simulation counts per move throughout training. Surprisingly, and in contrast to previous work \cite{negative_mcts_results}, even with only 6 simulations per move -- fewer than the number of actions -- \muzero{} learned an effective policy and improved rapidly. With more simulations performance jumped significantly higher. For analysis of the policy improvement during each individual iteration, see also Figure \ref{fig:extended_ablations} C and D.

\section{Conclusions}

Many of the breakthroughs in artificial intelligence have been based on either high-performance planning \cite{campbell:deep-blue, Silver16AG, Silver18AZ} or model-free reinforcement learning methods \cite{dqn, openai_dota, alphastar}. In this paper we have introduced a method that combines the benefits of both approaches. Our algorithm, \muzero{}, has both matched the superhuman performance of high-performance planning algorithms in their favored domains -- logically complex board games such as chess and Go -- and outperformed state-of-the-art model-free RL algorithms in their favored domains -- visually complex Atari games. Crucially, our method does not require any knowledge of the game rules or environment dynamics, potentially paving the way towards the application of powerful learning and planning methods to a host of real-world domains for which there exists no perfect simulator.

\section{Acknowledgments}

Lorrayne Bennett, Oliver Smith and Chris Apps for organizational assistance; Koray Kavukcuoglu for reviewing the paper; Thomas Anthony, Matthew Lai, Nenad Tomasev, Ulrich Paquet, Sumedh Ghaisas for many fruitful discussions; and the rest of the DeepMind team for their support.

\bibliographystyle{plain}
\bibliography{main}

\setcounter{table}{0}
\renewcommand{\thetable}{S\arabic{table}}%
\setcounter{figure}{0}
\renewcommand{\thefigure}{S\arabic{figure}}%

\subsection*{Supplementary Materials}

\begin{itemize}
\item Pseudocode description of the \muzero{} algorithm.
\item Data for Figures \ref{fig:results}, \ref{fig:ablations}, \ref{fig:muzero_equations}, \ref{fig:extended_ablations}, \ref{fig:learning_curve_atari} and Tables \ref{tab:atari-comparison}, \ref{tab:atari-results-at30}, \ref{tab:atari-results-at30-rnd-starts} in JSON format.
\end{itemize}

Supplementary materials can be accessed from the ancillary file section of the arXiv submission.

\begin{appendices}
\label{methods}

\section{Comparison to \alphazero}

\muzero{} is designed for a more general setting than  \alphagozero{} \cite{Silver17AG0} and \alphazero{} \cite{Silver18AZ}.

In \alphagozero{}  and \alphazero{} the planning process makes use of two separate components: a simulator implements the rules of the game, which are used to update the state of the game while traversing the search tree; and a neural network jointly predicts the corresponding policy and value of a board position produced by the simulator (see Figure \ref{fig:recurrent_search} A).

Specifically, \alphagozero{} and \alphazero{} use knowledge of the rules of the game in three places: (1) state transitions in the search tree, (2) actions available at each node of the search tree, (3) episode termination within the search tree.  In \muzero{}, all of these have been replaced with the use of a single implicit model learned by a neural network (see Figure \ref{fig:recurrent_search} B):

\begin{enumerate}[1)]
\item State transitions. \alphazero{} had access to a perfect simulator of the true dynamics process. In contrast, \muzero{} employs a learned dynamics model within its search. Under this model, each node in the tree is represented by a corresponding hidden state; by providing a hidden state $s_{k-1}$ and an action $a_k$ to the model the search algorithm can transition to a new node $s_{k}$ = $g(s_{k-1}, a_k)$.
\item Actions available. \alphazero{} used the set of legal actions obtained from the simulator to mask the prior produced by the network everywhere in the search tree. \muzero{} only masks legal actions at the root of the search tree where the environment can be queried, but does not perform any masking within the search tree. This is possible because the network rapidly learns not to predict actions that never occur in the trajectories it is trained on.
\item Terminal nodes. \alphazero{} stopped the search at tree nodes representing terminal states and used the terminal value provided by the simulator instead of the value produced by the network. \muzero{} does not give special treatment to terminal nodes and always uses the value predicted by the network. Inside the tree, the search can proceed past a terminal node - in this case the network is expected to always predict the same value. This is achieved by treating terminal states as absorbing states during training.
\end{enumerate}

In addition, \muzero{} is designed to operate in the general reinforcement learning setting: single-agent domains with discounted intermediate rewards of arbitrary magnitude. In contrast, \alphagozero{} and \alphazero{} were designed to operate in two-player games with undiscounted terminal rewards of $\pm 1$.

\section{Search}

We now describe the search algorithm used by \muzero{}. Our approach is based upon Monte-Carlo tree search with upper confidence bounds, an approach to planning that converges asymptotically to the optimal policy in single agent domains and to the minimax value function in zero sum games \cite{kocsis-sepesvari}.

Every node of the search tree is associated with an internal state $s$.
For each action $a$ from $s$ there is an edge $(s,a)$ that stores a set of statistics
$\{N(s,a), Q(s,a), P(s,a), R(s,a), S(s,a) \}$, respectively representing visit counts $N$,
mean value $Q$, policy $P$, reward $R$, and state transition $S$.

Similar to \alphazero{}, the search is divided into three stages, repeated for a number of simulations.

{\bf Selection}: Each simulation starts from the internal root state $s^0$, and finishes when the simulation reaches a leaf node $s^l$.
For each hypothetical time-step $k = 1 ... l$ of the simulation, an action $a^k$ is selected according to the stored statistics for
internal state $s^{k-1}$, by maximizing over an upper confidence bound \cite{rosin:puct}\cite{Silver18AZ},

\begin{equation} \label{pUCT}
a^{k} = \arg\max_{a}\bigg[
Q(s, a) + P(s, a) \cdot \frac{\sqrt{\sum_b N(s, b)}}{1 + N(s, a)} \bigg(c_1 + \log\Big(\frac{\sum_b N(s, b) + c_2 + 1}{c_2}\Big)\bigg) \bigg]
\end{equation}

The constants $c_1$ and $c_2$ are used to control the influence of the prior $P(s, a)$ relative to the value $Q(s, a)$ as nodes are visited more often. In our experiments, $c_1 = 1.25$ and $c_2 = 19652$.

For $k<l$, the next state and reward are looked up in the state transition and reward table $s^{k} = S(s^{k-1}, a^{k})$, $r^{k} = R(s^{k-1}, a^{k})$.

{\bf Expansion}: At the final time-step $l$ of the simulation, the reward
and state are computed by the dynamics function, $r^l, s^l = g_\theta(s^{l-1}, a^{l})$,
and stored in the corresponding tables, $R(s^{l-1}, a^{l}) = r^l, S(s^{l-1}, a^{l}) = s^l$.
The policy and value are computed by the prediction function, $\mathbf{p}^l, v^l = f_\theta(s^l)$.
A new node, corresponding to state $s^l$ is added to the search tree.
Each edge $(s^l, a)$ from the newly expanded node is initialized to
$\{ N(s^l, a)=0, Q(s^l, a)=0, P(s^l, a)=\mathbf{p}^l \}$.
Note that the search algorithm makes at most one call to the dynamics function and prediction function respectively per simulation; the computational cost is of the same order as in \alphazero{}.

{\bf Backup}: At the end of the simulation, the statistics along the trajectory are updated.
The backup is generalized to the case where the environment can emit intermediate rewards, have a discount $\gamma$ different from $1$,
and the value estimates are unbounded \footnote{In board games the discount is assumed to be $1$ and there are no intermediate rewards.}.
For $k= l ... 0$, we form an $l-k$-step estimate of the cumulative discounted reward, bootstrapping
from the value function $v^l$,

\begin{equation}
G^k = \sum_{\tau=0}^{l-1-k}{\gamma ^ \tau r_{k+ 1 +\tau}} + \gamma^{l - k} v^l
\end{equation}

For $k= l ... 1$, we update the statistics for each edge $(s^{k-1}, a^{k})$ in the simulation path as follows,

\begin{equation}
\begin{aligned}
Q(s^{k-1},a^k) & := \frac{N(s^{k-1},a^k) \cdot Q(s^{k-1},a^k) + G^k}{N(s^{k-1},a^k) + 1} \\
N(s^{k-1},a^k) & := N(s^{k-1},a^k) + 1
\end{aligned}
\end{equation}

In two-player zero sum games the value functions are assumed to be bounded within the $[0, 1]$ interval. This choice allows us to
combine value estimates with probabilities using the pUCT rule (Eqn \ref{pUCT}).
However, since in many environments the value is unbounded, it is necessary
to adjust the pUCT rule. A simple solution would be to use the maximum score
that can be observed in the environment to either re-scale the value or set the
pUCT constants appropriately \cite{MCTS_single_agent}. However, both solutions are
game specific and require adding prior knowledge to the \muzero{} algorithm. To avoid this, \muzero{} computes normalized $Q$ value estimates $\overline{Q} \in [0, 1]$
by using the minimum-maximum values observed in the search tree up to that point.
When a node is reached during the selection stage, the algorithm computes the normalized $\overline{Q}$ values of its edges to be used in the pUCT rule using the equation below:

\begin{equation}
\overline{Q}(s^{k-1}, a^k) = \frac{Q(s^{k-1}, a^k) - \min_{s, a \in Tree}Q(s, a)}{\max_{s,a \in Tree} Q(s, a) - \min_{s,a \in Tree} Q(s, a)}
\end{equation}

\section{Hyperparameters}

For simplicity we preferentially use the same architectural choices and hyperparameters as in previous work. Specifically, we started with the network architecture and search choices of \alphazero{} \cite{Silver18AZ}. For board games, we use the same UCB constants, dirichlet exploration noise and the same 800 simulations per search as in \alphazero{}.

Due to the much smaller branching factor and simpler policies in Atari, we only used 50 simulations per search to speed up experiments. As shown in Figure \ref{fig:ablations}B, the algorithm is not very sensitive to this choice. We also use the same discount ($0.997$) and value transformation (see Network Architecture section) as R2D2 \cite{r2d2}.

For parameter values not mentioned in the text, please refer to the pseudocode.

\section{Data Generation}

To generate training data, the latest checkpoint of the network (updated every 1000 training steps) is used to play games with MCTS. In the board games Go, chess and shogi the search is run for 800 simulations per move to pick an action; in Atari due to the much smaller action space 50 simulations per move are sufficient.

For board games, games are sent to the training job as soon as they finish. Due to the much larger length of Atari games (up to 30 minutes or 108,000 frames), intermediate sequences are sent every 200 moves. In board games, the training job keeps an in-memory replay buffer of the most recent 1 million games received; in Atari, where the visual observations are larger, the most recent 125 thousand sequences of length 200 are kept.

During the generation of experience in the board game domains, the same exploration scheme as the one described in
\alphazero{} \cite{Silver18AZ} is used. Using a variation of this scheme, in the Atari domain actions are sampled from the visit count distribution throughout the duration of each game, instead of just the first $k$ moves. Moreover, the visit count distribution is parametrized using a temperature parameter $T$:

\begin{equation}
p_{\alpha} = \frac{N(\alpha)^{1 / T}}{\sum_{b} N(b)^{1 / T}}
\end{equation}

$T$ is decayed as a function of the number of training steps of the network. Specifically, for the first 500k training steps a temperature of $1$ is used, for the next 250k steps a temperature of $0.5$ and for the remaining 250k a temperature of $0.25$. This ensures that the action selection becomes greedier as training progresses.

\section{Network Input}

\subsection*{Representation Function}

The history over board states used as input to the representation function for Go, chess and shogi is represented similarly to \alphazero{} \cite{Silver18AZ}. In Go and shogi we encode the last 8 board states as in \alphazero{}; in chess we increased the history to the last 100 board states to allow correct prediction of draws.

For Atari, the input of the representation function includes the last 32 RGB frames at resolution 96x96 along with the last 32 actions that led to each of those frames. We encode the historical actions because unlike board games, an action in Atari does not necessarily have a visible effect on the observation. RGB frames are encoded as one plane per color, rescaled to the range $[0, 1]$, for red, green and blue respectively. We perform no other normalization, whitening or other preprocessing of the RGB input. Historical actions are encoded as simple bias planes, scaled as $a / 18$ (there are 18 total actions in Atari).

\subsection*{Dynamics Function}

The input to the dynamics function is the hidden state produced by the representation function or previous application of the dynamics function, concatenated with a representation of the action for the transition. Actions are encoded spatially in planes of the same resolution as the hidden state. In Atari, this resolution is 6x6 (see description of downsampling in Network Architecture section), in board games this is the same as the board size (19x19 for Go, 8x8 for chess, 9x9 for shogi).

In Go, a normal action (playing a stone on the board) is encoded as an all zero plane, with a single one in the position of the played stone. A pass is encoded as an all zero plane.

In chess, 8 planes are used to encode the action. The first one-hot plane encodes which position the piece was moved from. The next two planes encode which position the piece was moved to: a one-hot plane to encode the target position, if on the board, and a second binary plane to indicate whether the target was valid (on the board) or not. This is necessary because for simplicity our policy action space enumerates a superset of all possible actions, not all of which are legal, and we use the same action space for policy prediction and to encode the dynamics function input. The remaining five binary planes are used to indicate the type of promotion, if any (queen, knight, bishop, rook, none).

The encoding for shogi is similar, with a total of 11 planes. We use the first 8 planes to indicate where the piece moved from - either a board position (first one-hot plane) or the drop of one of the seven types of prisoner (remaining 7 binary planes). The next two planes are used to encode the target as in chess. The remaining binary plane indicates whether the move was a promotion or not.

In Atari, an action is encoded as a one hot vector which is tiled appropriately into planes.

\section{Network Architecture}

\label{sec:network_architecture}

The prediction function $\mathbf{p}^k, v^k = f_\theta(s^k)$ uses the same architecture as \alphazero{}: one or two convolutional layers that preserve the resolution but reduce the number of planes, followed by a fully connected layer to the size of the output.

For value and reward prediction in Atari we follow \cite{pohlen2018observe} in scaling targets using an invertible transform $h(x) = sign(x)(\sqrt{|x| + 1} - 1 + \epsilon x)$, where $\epsilon = 0.001$ in all our experiments. We then apply a transformation $\phi$ to the scalar reward and value targets in order to obtain equivalent categorical representations. We use a discrete support set of size $601$ with one support for every integer between $-300$ and $300$. Under this transformation, each scalar is represented as the linear combination of its two adjacent supports, such that the original value can be recovered by $x = x_{low} * p_{low} + x_{high} * p_{high}$. As an example, a target of $3.7$ would be represented as a weight of $0.3$ on the support for $3$ and a weight of $0.7$ on the support for $4$. The value and reward outputs of the network are also modeled using a softmax output of size $601$. During inference the actual value and rewards are obtained by first computing their expected value under their respective softmax distribution and subsequently by inverting the scaling transformation. Scaling and transformation of the value and reward happens transparently on the network side and is not visible to the rest of the algorithm.

Both the representation and dynamics function use the same architecture as \alphazero{}, but with 16 instead of 20 residual blocks \cite{he:resnet}. We use 3x3 kernels and 256 hidden planes for each convolution.

For Atari, where observations have large spatial resolution, the representation function starts with a sequence of convolutions with stride 2 to reduce the spatial resolution. Specifically, starting with an input observation of resolution 96x96 and 128 planes (32 history frames of 3 color channels each, concatenated with the corresponding 32 actions broadcast to planes), we downsample as follows:

\begin{itemize}
\item 1 convolution with stride 2 and 128 output planes, output resolution 48x48.
\item 2 residual blocks with 128 planes
\item 1 convolution with stride 2 and 256 output planes, output resolution 24x24.
\item 3 residual blocks with 256 planes.
\item Average pooling with stride 2, output resolution 12x12.
\item 3 residual blocks with 256 planes.
\item Average pooling with stride 2, output resolution 6x6.
\end{itemize}

The kernel size is 3x3 for all operations.

For the dynamics function (which always operates at the downsampled resolution of 6x6), the action is first encoded as an image, then stacked with the hidden state of the previous step along the plane dimension.

\section{Training}

During training, the \muzero{} network is unrolled for $K$ hypothetical steps and aligned to sequences sampled from the trajectories generated by the MCTS actors. Sequences are selected by sampling a state from any game in the replay buffer, then unrolling for $K$ steps from that state. In Atari, samples are drawn according to prioritized replay \cite{Schaul2016}, with priority $P(i) = \frac{p_i^\alpha}{\sum_k{p_k^\alpha}}$, where $p_i = |\nu_i - z_i|$, $\nu$ is the search value and $z$ the observed n-step return. To correct for sampling bias introduced by the prioritized sampling, we scale the loss using the importance sampling ratio $w_i = (\frac{1}{N} \cdot \frac{1}{P(i)})^\beta$. In all our experiments, we set $\alpha = \beta = 1$. For board games, states are sampled uniformly.

Each observation $o_t$ along the sequence also has a corresponding MCTS policy $\pi_t$, estimated value $\nu_t$ and environment reward $u_t$. At each unrolled step $k$ the network has a loss to the value, policy and reward target for that step, summed to produce the total loss for the \muzero{} network (see Equation \ref{muzero_eqn}). Note that, in board games without intermediate rewards, we omit the reward prediction loss. For board games, we bootstrap directly to the end of the game, equivalent to predicting the final outcome; for Atari we bootstrap for $n=10$ steps into the future.

To maintain roughly similar magnitude of gradient across different unroll steps, we scale the gradient in two separate locations:

\begin{itemize}
\item We scale the loss of each head by $\frac{1}{K}$, where $K$ is the number
      of unroll steps. This ensures that the total gradient has similar magnitude
      irrespective of how many steps we unroll for.
\item We also scale the gradient at the start of the dynamics function by
      $\frac{1}{2}$. This ensures that the total gradient applied to the
      dynamics function stays constant.
\end{itemize}

In the experiments reported in this paper, we always unroll for $K=5$ steps. For a detailed illustration, see Figure \ref{fig:recurrent_search}.

To improve the learning process and bound the activations, we also scale the hidden state to the same range as the action input ($[0, 1]$): $s_{scaled} = \frac{s - \min(s)}{\max(s) - \min(s)}$.

All experiments were run using third generation Google Cloud TPUs \cite{tpuv3}. For each board game, we used 16 TPUs for training and 1000 TPUs for selfplay. %
For each game in Atari, we used 8 TPUs for training and 32 TPUs for selfplay. %
The much smaller proportion of TPUs used for acting in Atari is due to the smaller number of simulations per move (50 instead of 800) and the smaller size of the dynamics function compared to the representation function.

\section{Reanalyze}
\label{reanalyze}

To improve the sample efficiency of \muzero{} we introduced a second variant of the algorithm, \reanalyze{}. \reanalyze{} revisits its past time-steps and re-executes its search using the latest model parameters, potentially resulting in a better quality policy than the original search. This fresh policy is used as the policy target for 80\% of updates during \muzero{} training. Furthermore, a target network \cite{dqn} $\cdot, v^- = f_{\theta^-}(s^0)$, based on recent parameters $\theta^-$, is used to provide a fresher, stable $n$-step bootstrapped target for the value function, $z_t = u_{t+1} + \gamma u_{t+2} + ... + \gamma^{n-1} u_{t+n} + \gamma^n v^-_{t+n}$. In addition, several other hyperparameters were adjusted -- primarily to increase sample reuse and avoid overfitting of the value function. Specifically, 2.0 samples were drawn per state, instead of 0.1; the value target was weighted down to 0.25 compared to weights of 1.0 for policy and reward targets; and the $n$-step return was reduced to $n=5$ steps instead of $n=10$ steps.

\section{Evaluation}

We evaluated the relative strength of \muzero{} (Figure \ref{fig:results}) in board games by measuring the Elo rating of each player. We estimate the probability that player $a$ will defeat player $b$ by a logistic function $p(a \text{ defeats } b) = (1 + 10^{(c_{\mathrm{elo}} (e(b) - e(a)))})^{-1}$, and estimate the ratings $e(\cdot)$ by Bayesian logistic regression, computed by the \emph{BayesElo} program \cite{coulom:bayeselo} using the standard constant $c_{\mathrm{elo}} = 1/400$.

Elo ratings were computed from the results of a 800 simulations per move tournament between iterations of \muzero{} during training, and also a baseline player: either \stockfish{}, \elmo{} or \alphazero{} respectively. Baseline players used an equivalent search time of 100ms per move. The Elo rating of the baseline players was anchored to publicly available values~\cite{Silver18AZ}.

In Atari, we computed mean reward over 1000 episodes per game, limited to the standard 30 minutes or 108,000 frames per episode \cite{gorila}, using 50 simulations per move unless indicated otherwise. In order to mitigate the effects of the deterministic nature of the Atari simulator we employed two different evaluation strategies: 30 noop random starts and human starts. For the former, at the beginning of each episode a random number of between 0 and 30 noop actions are applied to the simulator before handing control to the agent. For the latter, start positions are sampled from human expert play to initialize the Atari simulator before handing the control to the agent \cite{gorila}.

\end{appendices}

\begin{figure} %
\includegraphics[width=\textwidth]{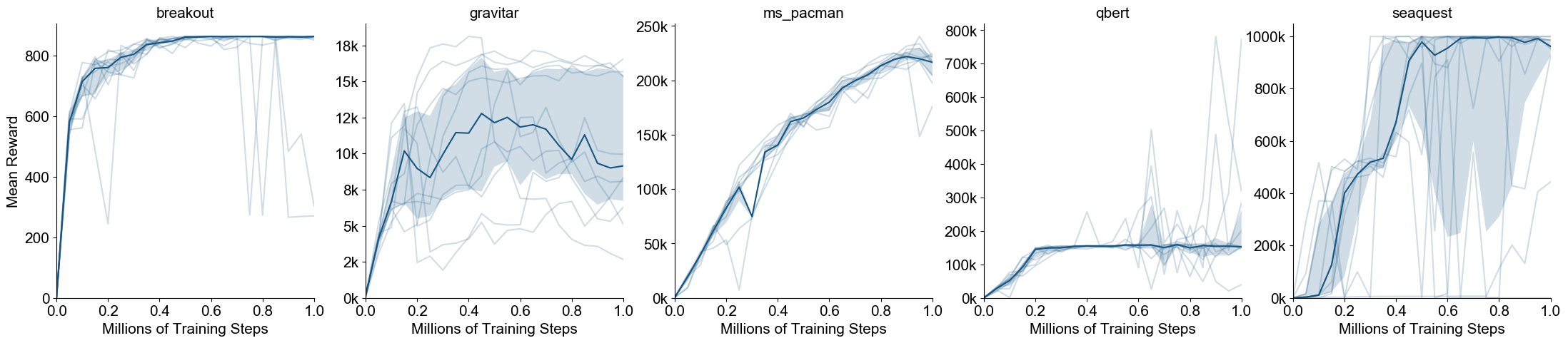}
\caption{
\label{fig:atari_repeatability}
\textbf{Repeatability of \muzero{} in Atari for five games.} Total reward is shown on the y-axis, millions of training steps on the x-axis. Dark line indicates median score across 10 separate training runs, light lines indicate individual training runs, and the shaded region indicates 25th to 75th percentile.
}
\end{figure} %

\begin{figure} %

\begin{align*}
&\text{Model} \\
&\left.
\begin{array}{ll}
s^0 &= h_\theta(o_1, ..., o_t) \\
r^k, s^k &= g_\theta(s^{k-1}, a^k) \\
\mathbf{p}^k, v^k &= f_\theta(s^k)
\end{array}
\right\} \;\;
\mathbf{p}^k, v^k, r^k = \mu_\theta(o_1, ..., o_t, a^1, ..., a^k)
\end{align*}

\begin{align*}
&\text{Search}\\
\nu_t, \pi_t &= MCTS(s^0_t, \mu_\theta) \\
a_t &\sim \pi_t \\
\\
&\text{Learning Rule} \\
\mathbf{p}^k_t, v^k_t, r^k_t &= \mu_\theta(o_1, ..., o_t, a_{t+1}, ..., a_{t+k}) \\
z_t &= \left\{
\begin{array}{lr}
u_T
& \text{ for games } \\
u_{t+1} + \gamma u_{t+2} + ... + \gamma^{n-1} u_{t+n} + \gamma^n \nu_{t+n}
& \text{ for general MDPs }
\end{array}
\right. \\
l_t(\theta) &= \sum_{k=0}^K l^r (u_{t+k}, r_t^k) + l^v(z_{t+k}, v^k_t) + l^p(\pi_{t+k}, p^k_t)  + c ||\theta||^2 \\
\\
&\text{Losses} \\
l^r(u, r) &= \left\{
\begin{array}{lr}
0 & \text{ for games } \\
\boldsymbol{\phi}(u)^T \log \mathbf{r} & \text{ for general MDPs }
\end{array}
\right. \\
l^v(z, q) &= \left\{
\begin{array}{lr}
(z - q)^2 & \text{ for games } \\
\boldsymbol{\phi}(z)^T \log \mathbf{q} & \text{ for general MDPs }
\end{array}
\right. \\
l^p(\pi, p) &= \boldsymbol{\pi}^T \log \mathbf{p}
\end{align*}

\caption{
\label{fig:muzero_equations}
\textbf{Equations summarising the \muzero{} algorithm}. Here, $\phi(x)$ refers to the representation of a real number $x$ through a linear combination of its adjacent integers, as described in the Network Architecture section.
}
\end{figure} %

\begin{figure} %
\includegraphics[width=0.5\textwidth]{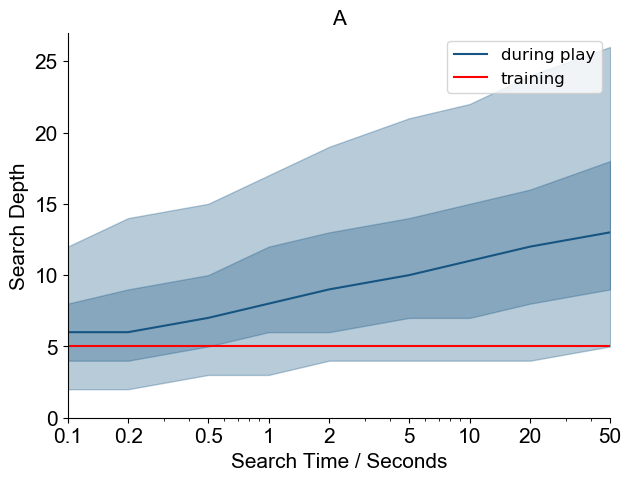}
\includegraphics[width=0.5\textwidth]{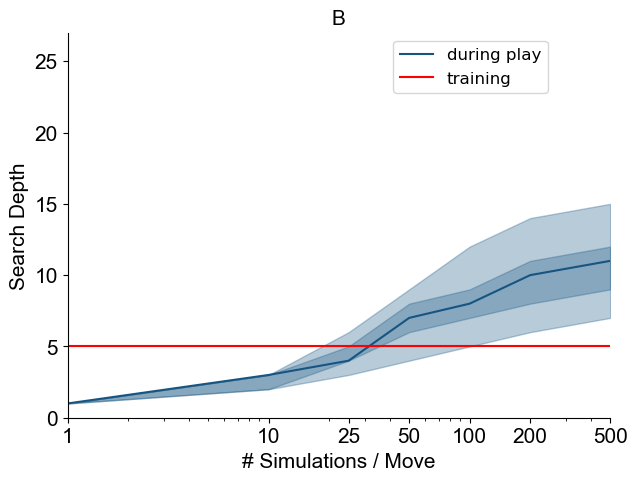}
\includegraphics[width=0.5\textwidth]{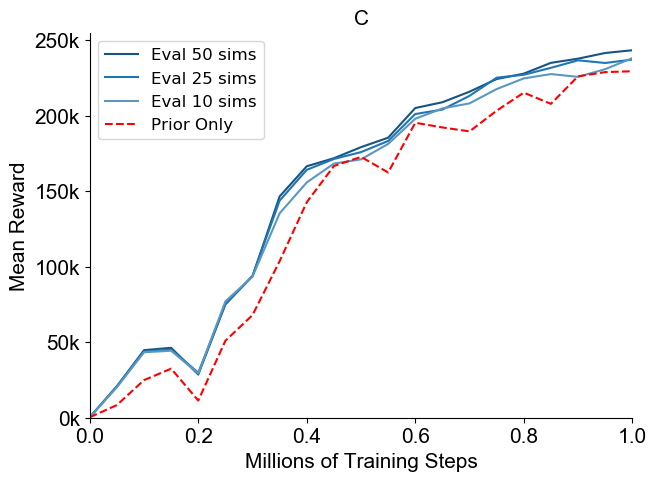}
\includegraphics[width=0.5\textwidth]{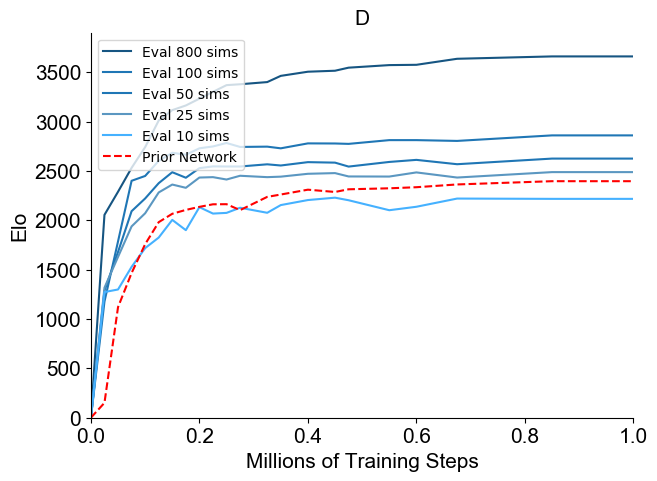}
\caption{
\label{fig:extended_ablations}
\textbf{Details of \muzero{} evaluations (A-B) and policy improvement ablations (C-D)}.
(\textbf{A-B}) Distribution of evaluation depth in the search tree for the learned model for the evaluations in Figure \ref{fig:ablations}A-B. The network was trained over 5 hypothetical steps, as indicated by the red line. Dark blue line indicates median depth from the root, dark shaded region shows 25th to 75th percentile, light shaded region shows 5th to 95th percentile.
(\textbf{C}) Policy improvement in Ms. Pacman - a single network was trained at 50 simulations per search and is evaluated at different numbers of simulations per search, including playing according to the argmax of the raw policy network. The policy improvement effect of the search over the raw policy network is clearly visible throughout training. This consistent gap between the performance with and without search highlights the policy improvement that \muzero{} exploits, by continually updating towards the improved policy, to efficiently progress towards the optimal policy.
(\textbf{D}) Policy improvement in Go - a single network was trained at 800 simulations per search and is evaluated at different numbers of simulations per search. In Go, the playing strength improvement from longer searches is much larger than in Ms.~Pacman and persists throughout training, consistent with previous results in \cite{Silver17AG0}. This suggests, as might intuitively be expected, that the benefit of models is greatest in precision planning domains.
}
\end{figure} %

\begin{table}
\scriptsize

\begin{center}\begin{tabularx}{0.900000\textwidth}{X| r r r r r r r}
\toprule
Game & Random & Human & SimPLe \cite{kaiser:simple} & Ape-X \cite{apex} & R2D2 \cite{r2d2} & \muzero{} & \muzero{} normalized  \\
\midrule
alien &227.75 & 7,127.80 & 616.90 & 40,805.00 & 229,496.90 & \textbf{741,812.63} & 10,747.5 \%\\
amidar &5.77 & 1,719.53 & 74.30 & 8,659.00 & \textbf{29,321.40} & 28,634.39 & 1,670.5 \%\\
assault &222.39 & 742.00 & 527.20 & 24,559.00 & 108,197.00 & \textbf{143,972.03} & 27,664.9 \%\\
asterix &210.00 & 8,503.33 & 1,128.30 & 313,305.00 & \textbf{999,153.30} & 998,425.00 & 12,036.4 \%\\
asteroids &719.10 & 47,388.67 & 793.60 & 155,495.00 & 357,867.70 & \textbf{678,558.64} & 1,452.4 \%\\
atlantis &12,850.00 & 29,028.13 & 20,992.50 & 944,498.00 & 1,620,764.00 & \textbf{1,674,767.20} & 10,272.6 \%\\
bank heist &14.20 & 753.13 & 34.20 & 1,716.00 & \textbf{24,235.90} & 1,278.98 & 171.2 \%\\
battle zone &2,360.00 & 37,187.50 & 4,031.20 & 98,895.00 & 751,880.00 & \textbf{848,623.00} & 2,429.9 \%\\
beam rider &363.88 & 16,926.53 & 621.60 & 63,305.00 & 188,257.40 & \textbf{454,993.53} & 2,744.9 \%\\
berzerk &123.65 & 2,630.42 & - & 57,197.00 & 53,318.70 & \textbf{85,932.60} & 3,423.1 \%\\
bowling &23.11 & 160.73 & 30.00 & 18.00 & 219.50 & \textbf{260.13} & 172.2 \%\\
boxing &0.05 & 12.06 & 7.80 & \textbf{100.00} & 98.50 & \textbf{100.00} & 832.2 \%\\
breakout &1.72 & 30.47 & 16.40 & 801.00 & 837.70 & \textbf{864.00} & 2,999.2 \%\\
centipede &2,090.87 & 12,017.04 & - & 12,974.00 & 599,140.30 & \textbf{1,159,049.27} & 11,655.6 \%\\
chopper command &811.00 & 7,387.80 & 979.40 & 721,851.00 & 986,652.00 & \textbf{991,039.70} & 15,056.4 \%\\
crazy climber &10,780.50 & 35,829.41 & 62,583.60 & 320,426.00 & 366,690.70 & \textbf{458,315.40} & 1,786.6 \%\\
defender &2,874.50 & 18,688.89 & - & 411,944.00 & 665,792.00 & \textbf{839,642.95} & 5,291.2 \%\\
demon attack &152.07 & 1,971.00 & 208.10 & 133,086.00 & 140,002.30 & \textbf{143,964.26} & 7,906.4 \%\\
double dunk &-18.55 & -16.40 & - & \textbf{24.00} & 23.70 & 23.94 & 1,976.3 \%\\
enduro &0.00 & 860.53 & - & 2,177.00 & 2,372.70 & \textbf{2,382.44} & 276.9 \%\\
fishing derby &-91.71 & -38.80 & -90.70 & 44.00 & 85.80 & \textbf{91.16} & 345.6 \%\\
freeway &0.01 & 29.60 & 16.70 & \textbf{34.00} & 32.50 & 33.03 & 111.6 \%\\
frostbite &65.20 & 4,334.67 & 236.90 & 9,329.00 & 315,456.40 & \textbf{631,378.53} & 14,786.7 \%\\
gopher &257.60 & 2,412.50 & 596.80 & 120,501.00 & 124,776.30 & \textbf{130,345.58} & 6,036.8 \%\\
gravitar &173.00 & 3,351.43 & 173.40 & 1,599.00 & \textbf{15,680.70} & 6,682.70 & 204.8 \%\\
hero &1,026.97 & 30,826.38 & 2,656.60 & 31,656.00 & 39,537.10 & \textbf{49,244.11} & 161.8 \%\\
ice hockey &-11.15 & 0.88 & -11.60 & 33.00 & \textbf{79.30} & 67.04 & 650.0 \%\\
jamesbond &29.00 & 302.80 & 100.50 & 21,323.00 & 25,354.00 & \textbf{41,063.25} & 14,986.9 \%\\
kangaroo &52.00 & 3,035.00 & 51.20 & 1,416.00 & 14,130.70 & \textbf{16,763.60} & 560.2 \%\\
krull &1,598.05 & 2,665.53 & 2,204.80 & 11,741.00 & 218,448.10 & \textbf{269,358.27} & 25,083.4 \%\\
kung fu master &258.50 & 22,736.25 & 14,862.50 & 97,830.00 & \textbf{233,413.30} & 204,824.00 & 910.1 \%\\
montezuma revenge &0.00 & \textbf{4,753.33} & - & 2,500.00 & 2,061.30 & 0.00 & 0.0 \%\\
ms pacman &307.30 & 6,951.60 & 1,480.00 & 11,255.00 & 42,281.70 & \textbf{243,401.10} & 3,658.7 \%\\
name this game &2,292.35 & 8,049.00 & 2,420.70 & 25,783.00 & 58,182.70 & \textbf{157,177.85} & 2,690.5 \%\\
phoenix &761.40 & 7,242.60 & - & 224,491.00 & 864,020.00 & \textbf{955,137.84} & 14,725.3 \%\\
pitfall &-229.44 & \textbf{6,463.69} & - & -1.00 & 0.00 & 0.00 & 3.4 \%\\
pong &-20.71 & 14.59 & 12.80 & \textbf{21.00} & \textbf{21.00} & \textbf{21.00} & 118.2 \%\\
private eye &24.94 & \textbf{69,571.27} & 35.00 & 50.00 & 5,322.70 & 15,299.98 & 22.0 \%\\
qbert &163.88 & 13,455.00 & 1,288.80 & 302,391.00 & \textbf{408,850.00} & 72,276.00 & 542.6 \%\\
riverraid &1,338.50 & 17,118.00 & 1,957.80 & 63,864.00 & 45,632.10 & \textbf{323,417.18} & 2,041.1 \%\\
road runner &11.50 & 7,845.00 & 5,640.60 & 222,235.00 & 599,246.70 & \textbf{613,411.80} & 7,830.5 \%\\
robotank &2.16 & 11.94 & - & 74.00 & 100.40 & \textbf{131.13} & 1,318.7 \%\\
seaquest &68.40 & 42,054.71 & 683.30 & 392,952.00 & \textbf{999,996.70} & 999,976.52 & 2,381.5 \%\\
skiing &-17,098.09 & \textbf{-4,336.93} & - & -10,790.00 & -30,021.70 & -29,968.36 & -100.9 \%\\
solaris &1,236.30 & \textbf{12,326.67} & - & 2,893.00 & 3,787.20 & 56.62 & -10.6 \%\\
space invaders &148.03 & 1,668.67 & - & 54,681.00 & 43,223.40 & \textbf{74,335.30} & 4,878.7 \%\\
star gunner &664.00 & 10,250.00 & - & 434,343.00 & \textbf{717,344.00} & 549,271.70 & 5,723.0 \%\\
surround &-9.99 & 6.53 & - & 7.00 & 9.90 & \textbf{9.99} & 120.9 \%\\
tennis &-23.84 & -8.27 & - & \textbf{24.00} & -0.10 & 0.00 & 153.1 \%\\
time pilot &3,568.00 & 5,229.10 & - & 87,085.00 & 445,377.30 & \textbf{476,763.90} & 28,486.9 \%\\
tutankham &11.43 & 167.59 & - & 273.00 & 395.30 & \textbf{491.48} & 307.4 \%\\
up n down &533.40 & 11,693.23 & 3,350.30 & 401,884.00 & 589,226.90 & \textbf{715,545.61} & 6,407.0 \%\\
venture &0.00 & 1,187.50 & - & 1,813.00 & \textbf{1,970.70} & 0.40 & 0.0 \%\\
video pinball &0.00 & 17,667.90 & - & 565,163.00 & \textbf{999,383.20} & 981,791.88 & 5,556.9 \%\\
wizard of wor &563.50 & 4,756.52 & - & 46,204.00 & 144,362.70 & \textbf{197,126.00} & 4,687.9 \%\\
yars revenge &3,092.91 & 54,576.93 & 5,664.30 & 148,595.00 & \textbf{995,048.40} & 553,311.46 & 1,068.7 \%\\
zaxxon &32.50 & 9,173.30 & - & 42,286.00 & 224,910.70 & \textbf{725,853.90} & 7,940.5 \%\\
\midrule
\# best & 0 & 5 & 0 & 5 & 13 & 37\\
\bottomrule
\end{tabularx}
\end{center}

\caption{
\label{tab:atari-results-at30}
\textbf{Evaluation of \muzero{} in Atari for individual games with 30 random no-op starts.} Best result for each game highlighted in \textbf{bold}. Each episode is limited to a maximum of 30 minutes of game time (108k frames). SimPLe was only evaluated on 36 of the 57 games, unavailable results are indicated with `-'. Human normalized score is calculated as $s_{normalized} = \frac{s_{agent} - s_{random}}{s_{human} - s_{random}}$.
}
\end{table}

\begin{table}
\scriptsize

\begin{center}\begin{tabularx}{0.750000\textwidth}{X| r r r r r}
\toprule
Game & Random & Human & Ape-X \cite{apex} & \muzero{} & \muzero{} normalized  \\
\midrule
alien &128.30 & 6,371.30 & 17,732.00 & \textbf{713,387.37} & 11,424.9 \%\\
amidar &11.79 & 1,540.43 & 1,047.00 & \textbf{26,638.80} & 1,741.9 \%\\
assault &166.95 & 628.89 & 24,405.00 & \textbf{143,900.58} & 31,115.2 \%\\
asterix &164.50 & 7,536.00 & 283,180.00 & \textbf{985,801.95} & 13,370.9 \%\\
asteroids &877.10 & 36,517.30 & 117,303.00 & \textbf{606,971.12} & 1,700.6 \%\\
atlantis &13,463.00 & 26,575.00 & 918,715.00 & \textbf{1,653,202.50} & 12,505.6 \%\\
bank heist &21.70 & 644.50 & \textbf{1,201.00} & 962.11 & 151.0 \%\\
battle zone &3,560.00 & 33,030.00 & 92,275.00 & \textbf{791,387.00} & 2,673.3 \%\\
beam rider &254.56 & 14,961.02 & 72,234.00 & \textbf{419,460.76} & 2,850.5 \%\\
berzerk &196.10 & 2,237.50 & 55,599.00 & \textbf{87,308.60} & 4,267.3 \%\\
bowling &35.16 & 146.46 & 30.00 & \textbf{194.03} & 142.7 \%\\
boxing &-1.46 & 9.61 & \textbf{81.00} & 56.60 & 524.5 \%\\
breakout &1.77 & 27.86 & 757.00 & \textbf{849.59} & 3,249.6 \%\\
centipede &1,925.45 & 10,321.89 & 5,712.00 & \textbf{1,138,294.60} & 13,533.9 \%\\
chopper command &644.00 & 8,930.00 & 576,602.00 & \textbf{932,370.10} & 11,244.6 \%\\
crazy climber &9,337.00 & 32,667.00 & 263,954.00 & \textbf{412,213.90} & 1,726.9 \%\\
defender &1,965.50 & 14,296.00 & 399,865.00 & \textbf{823,636.00} & 6,663.7 \%\\
demon attack &208.25 & 3,442.85 & 133,002.00 & \textbf{143,858.05} & 4,441.0 \%\\
double dunk &-15.97 & -14.37 & 22.00 & \textbf{23.12} & 2,443.1 \%\\
enduro &-81.84 & 740.17 & 2,042.00 & \textbf{2,264.20} & 285.4 \%\\
fishing derby &-77.11 & 5.09 & 22.00 & \textbf{57.45} & 163.7 \%\\
freeway &0.17 & 25.61 & \textbf{29.00} & 28.38 & 110.9 \%\\
frostbite &90.80 & 4,202.80 & 6,512.00 & \textbf{613,944.04} & 14,928.3 \%\\
gopher &250.00 & 2,311.00 & 121,168.00 & \textbf{129,218.68} & 6,257.6 \%\\
gravitar &245.50 & 3,116.00 & 662.00 & \textbf{3,390.65} & 109.6 \%\\
hero &1,580.30 & 25,839.40 & 26,345.00 & \textbf{44,129.55} & 175.4 \%\\
ice hockey &-9.67 & 0.53 & 24.00 & \textbf{52.40} & 608.5 \%\\
jamesbond &33.50 & 368.50 & 18,992.00 & \textbf{39,107.20} & 11,663.8 \%\\
kangaroo &100.00 & 2,739.00 & 578.00 & \textbf{13,210.50} & 496.8 \%\\
krull &1,151.90 & 2,109.10 & 8,592.00 & \textbf{257,706.70} & 26,802.6 \%\\
kung fu master &304.00 & 20,786.80 & 72,068.00 & \textbf{174,623.60} & 851.1 \%\\
montezuma revenge &25.00 & \textbf{4,182.00} & 1,079.00 & 57.10 & 0.8 \%\\
ms pacman &197.80 & 15,375.05 & 6,135.00 & \textbf{230,650.24} & 1,518.4 \%\\
name this game &1,747.80 & 6,796.00 & 23,830.00 & \textbf{152,723.62} & 2,990.7 \%\\
phoenix &1,134.40 & 6,686.20 & 188,789.00 & \textbf{943,255.07} & 16,969.6 \%\\
pitfall &-348.80 & \textbf{5,998.91} & -273.00 & -801.10 & -7.1 \%\\
pong &-17.95 & 15.46 & 19.00 & \textbf{19.20} & 111.2 \%\\
private eye &662.78 & \textbf{64,169.07} & 865.00 & 5,067.59 & 6.9 \%\\
qbert &159.38 & 12,085.00 & \textbf{380,152.00} & 39,302.10 & 328.2 \%\\
riverraid &588.30 & 14,382.20 & 49,983.00 & \textbf{315,353.33} & 2,281.9 \%\\
road runner &200.00 & 6,878.00 & 127,112.00 & \textbf{580,445.00} & 8,688.9 \%\\
robotank &2.42 & 8.94 & 69.00 & \textbf{128.80} & 1,938.3 \%\\
seaquest &215.50 & 40,425.80 & 377,180.00 & \textbf{997,601.01} & 2,480.4 \%\\
skiing &-15,287.35 & \textbf{-3,686.58} & -11,359.00 & -29,400.75 & -121.7 \%\\
solaris &2,047.20 & \textbf{11,032.60} & 3,116.00 & 2,108.08 & 0.7 \%\\
space invaders &182.55 & 1,464.90 & 50,699.00 & \textbf{57,450.41} & 4,465.9 \%\\
star gunner &697.00 & 9,528.00 & 432,958.00 & \textbf{539,342.70} & 6,099.5 \%\\
surround &-9.72 & 5.37 & 6.00 & \textbf{8.46} & 120.5 \%\\
tennis &-21.43 & -6.69 & \textbf{23.00} & -2.30 & 129.8 \%\\
time pilot &3,273.00 & 5,650.00 & 71,543.00 & \textbf{405,829.30} & 16,935.5 \%\\
tutankham &12.74 & 138.30 & 128.00 & \textbf{351.76} & 270.0 \%\\
up n down &707.20 & 9,896.10 & 347,912.00 & \textbf{607,807.85} & 6,606.9 \%\\
venture &18.00 & \textbf{1,039.00} & 936.00 & 21.10 & 0.3 \%\\
video pinball &0.00 & 15,641.09 & 873,989.00 & \textbf{970,881.10} & 6,207.2 \%\\
wizard of wor &804.00 & 4,556.00 & 46,897.00 & \textbf{196,279.20} & 5,209.9 \%\\
yars revenge &1,476.88 & 47,135.17 & 131,701.00 & \textbf{888,633.84} & 1,943.0 \%\\
zaxxon &475.00 & 8,443.00 & 37,672.00 & \textbf{592,238.70} & 7,426.8 \%\\
\midrule
\# best & 0 & 6 & 5 & 46\\
\bottomrule
\end{tabularx}
\end{center}

\caption{
\label{tab:atari-results-at30-rnd-starts}
\textbf{Evaluation of \muzero{} in Atari for individual games from human start positions.} Best result for each game highlighted in \textbf{bold}.  Each episode is limited to a maximum of 30 minutes of game time (108k frames).
}
\end{table}

\begin{figure}
\includegraphics[width=0.95\textwidth]{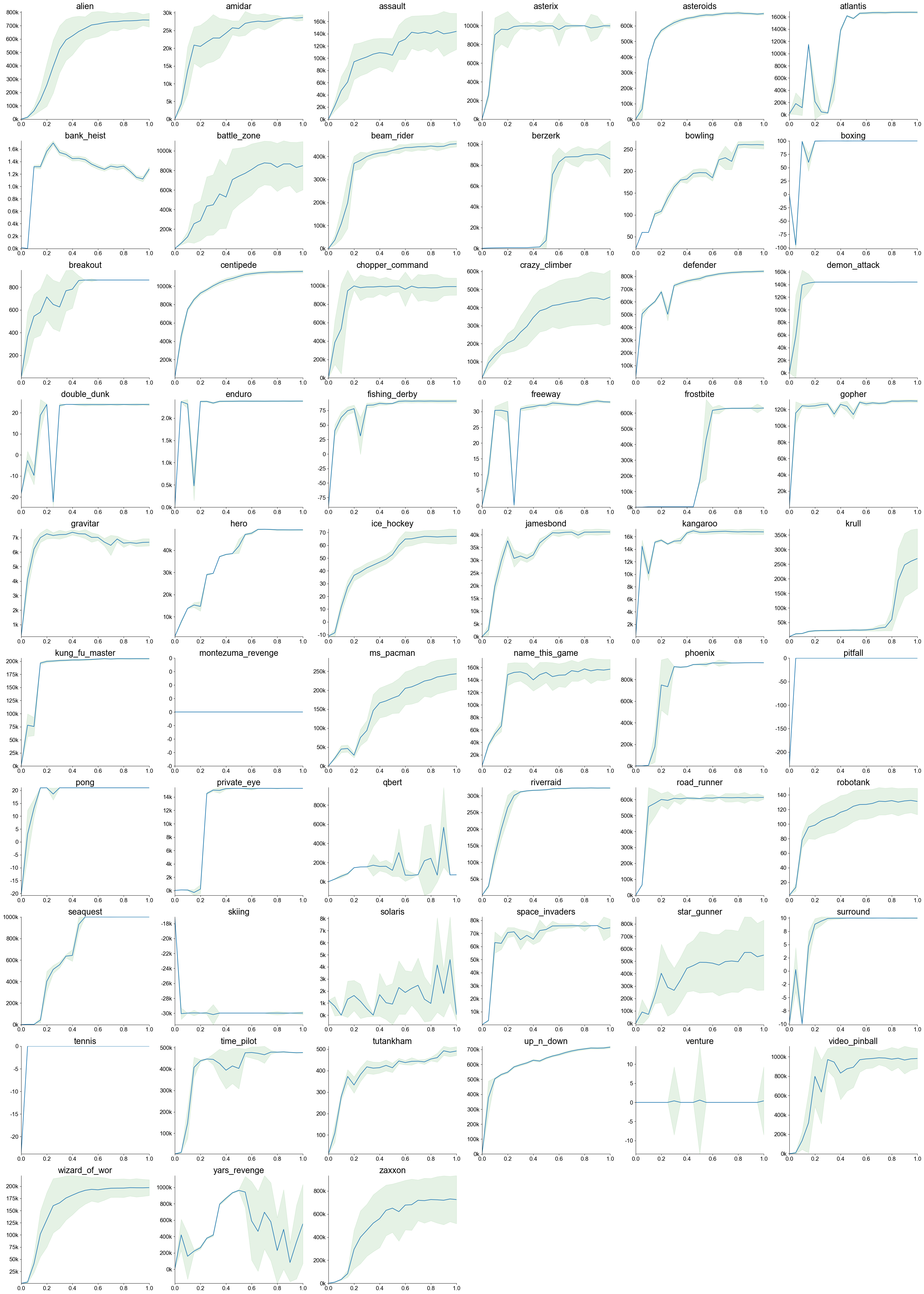}
\caption{
\label{fig:learning_curve_atari}
\textbf{Learning curves of \muzero{} in Atari for individual games.} Total reward is shown on the y-axis, millions of training steps on the x-axis. Line indicates mean score across 1000 evaluation games, shaded region indicates standard deviation.
}
\end{figure}

\end{document}